\pdfoutput=1

\documentclass{article} 
\usepackage{iclr2023_conference,times}

\iclrfinalcopy

\usepackage{amsmath,amsfonts,bm}

\def\eqref#1{equation~\ref{#1}}

\def\1{\bm{1}}

\DeclareMathAlphabet{\mathsfit}{\encodingdefault}{\sfdefault}{m}{sl}
\SetMathAlphabet{\mathsfit}{bold}{\encodingdefault}{\sfdefault}{bx}{n}

\DeclareMathOperator*{\argmax}{arg\,max}

\usepackage{hyperref,rotating}

\usepackage{latexsym}

\usepackage[normalem]{ulem}

\usepackage[T1]{fontenc}

\usepackage[utf8]{inputenc}

\usepackage{microtype}

\usepackage{natbib}
\usepackage{url}

\usepackage{rotating}  
\usepackage{float}
\usepackage{graphicx}
\usepackage{amsmath}

\usepackage{booktabs}
\usepackage{xspace,mfirstuc,tabulary}
\usepackage{multirow}
\usepackage{comment}
\usepackage{xcolor}

\usepackage{paralist}
\usepackage{longtable}

\usepackage{color}
\definecolor{myblue}{rgb}{0, 57, 230}
\definecolor{airforceblue}{rgb}{0.36, 0.54, 0.66}
\definecolor{navyblue}{rgb}{0, 0, 128}
\definecolor{ceruleanblue}{rgb}{0.16, 0.32, 0.75}
\definecolor{cornflowerblue}{rgb}{0.39, 0.58, 0.93}
\definecolor{denim}{rgb}{0.08, 0.38, 0.74}

\definecolor{azure(colorwheel)}{rgb}{0.0, 0.5, 1.0}
\definecolor{cornellred}{rgb}{0.7, 0.11, 0.11}

\definecolor{lemon}{rgb}{1.0, 0.97, 0.0}
\definecolor{amber(sae/ece)}{rgb}{1.0, 0.49, 0.0}
\definecolor{cadmiumorange}{rgb}{0.93, 0.53, 0.18}
\definecolor{darkorange}{rgb}{1.0, 0.55, 0.0}
\definecolor{debianred}{rgb}{0.84, 0.04, 0.33}
\definecolor{deepcarmine}{rgb}{0.66, 0.13, 0.24}
\definecolor{deepcarminepink}{rgb}{0.94, 0.19, 0.22}

\definecolor{brightgreen}{rgb}{0.4, 1.0, 0.0}
\definecolor{caribbeangreen}{rgb}{0.0, 0.8, 0.6}
\definecolor{chartreuse(web)}{rgb}{0.5, 1.0, 0.0}
\definecolor{darkpastelgreen}{rgb}{0.01, 0.75, 0.24}
\definecolor{electricgreen}{rgb}{0.0, 1.0, 0.0}
\definecolor{emerald}{rgb}{0.31, 0.78, 0.47}
\definecolor{cadmiumgreen}{rgb}{0.0, 0.42, 0.24}

\definecolor{darkmagenta}{rgb}{0.55, 0.0, 0.55}
\definecolor{darklavender}{rgb}{0.45, 0.31, 0.59}

\usepackage{amssymb}
\usepackage{amsfonts}
\usepackage{bbold}
\usepackage{bm}
\usepackage{float}
\usepackage{makecell}
\usepackage{subfigure}

\usepackage{caption}

\usepackage{bbm}

\usepackage{mathtools}

\DeclareSymbolFontAlphabet{\amsmathbb}{AMSb}%
\newcommand{\Y}[0]{\amsmathbb Y}
\newcommand{\BBC}[0]{\amsmathbb C}
\newcommand{\BBS}[0]{\amsmathbb S}

\newcommand{\EXP}[0]{\mathop{\amsmathbb E}}

\usepackage{pict2e,picture}
\makeatletter
\DeclareRobustCommand{\Arrow}[1][]{%
\check@mathfonts
\if\relax\detokenize{#1}\relax
\settowidth{\dimen@}{$\m@th\rightarrow$}%
\else
\setlength{\dimen@}{#1}%
\fi
\sbox\z@{\usefont{U}{lasy}{m}{n}\symbol{41}}%
\begin{picture}(\dimen@,\ht\z@)
\roundcap
\put(\dimexpr\dimen@-.7\wd\z@,0){\usebox\z@}
\put(0,\fontdimen22\textfont2){\line(1,0){\dimen@}}
\end{picture}%
}
\makeatother

\makeatletter
\newcommand{\thickhline}{%
    \noalign {\ifnum 0=`}\fi \hrule height 1pt
    \futurelet \reserved@a \@xhline
}
\makeatother

\makeatletter
\def\Cline#1#2{\@Cline#1#2\@nil}
\def\@Cline#1-#2#3\@nil{%
  \omit
  \@multicnt#1%
  \advance\@multispan\m@ne
  \ifnum\@multicnt=\@ne\@firstofone{&\omit}\fi
  \@multicnt#2%
  \advance\@multicnt-#1%
  \advance\@multispan\@ne
  \leaders\hrule\@height#3\hfill
  \cr}
\makeatother

\usepackage[bb=dsserif]{mathalpha}

\usepackage{array}
\newcolumntype{P}[1]{>{\centering\arraybackslash}p{#1}}
\definecolor{gblue}{RGB}{66,133,244}
\definecolor{gred}{RGB}{219,68,55}
\definecolor{gyellow}{RGB}{244,160,0}
\definecolor{ggreen}{RGB}{15,157,88}
\usepackage{float}
\usepackage{multicol}
\usepackage{blindtext}

\usepackage{lipsum}
\usepackage{enumitem}
\usepackage{tikz,pgfplots}

\usepackage{arydshln}
\usepackage{bibentry}

\usepackage{algorithm}
\usepackage{algpseudocode}
\algnewcommand\algorithmicbreak{\textbf{break}}
\algnewcommand\algorithmiccontinue{\textbf{continue}}

\usepackage{bold-extra}
\usepackage{wrapfig}

\usepackage{pifont}
\newcommand{\cmark}{\ding{51}}%
\newcommand{\xmark}{\ding{55}}%

\usepackage{ulem}

\usepackage{multicol}

\usepackage{pgfplots}
\usetikzlibrary{backgrounds}
\pgfplotsset{compat=1.12}

\usetikzlibrary{patterns}

\title{Text Summarization with Oracle Expectation}
\author{Yumo Xu \& Mirella Lapata\\
 Institute for Language, Cognition and Computation\\
 School of Informatics, University of Edinburgh\\
 10 Crichton Street, Edinburgh EH8 9AB\\
 \texttt{yumo.xu@ed.ac.uk}, 
 \texttt{mlap@inf.ed.ac.uk}}
\date{}

\begin{document}
\maketitle
\global\csname @topnum\endcsname 0
\global\csname @botnum\endcsname 0

\begin{abstract}
Extractive summarization produces summaries by identifying and concatenating the most important sentences in a document. 
Since most summarization datasets do not come with \textit{gold} labels indicating whether document sentences are summary-worthy, 
different labeling algorithms have been proposed to extrapolate \textit{oracle} extracts for model training. 
In this work, we identify two flaws with the widely used greedy labeling approach: it delivers suboptimal  and deterministic oracles. 
To alleviate both issues,  we propose a simple yet effective labeling algorithm that creates soft, expectation-based sentence labels. 
We define a new learning objective for extractive summarization 
which  incorporates  learning signals from multiple  oracle summaries and prove it  is equivalent to  estimating the oracle expectation for each
document sentence. Without any architectural modifications, the proposed labeling scheme achieves superior performance on a variety of summarization benchmarks across domains and languages, in both supervised and zero-shot settings.\footnote{Our code and
  models can be found at \url{https://github.com/yumoxu/oreo}.}

\end{abstract}

\section{Introduction}
Summarization is the process of condensing a source text into a
shorter version while preserving its information content. Thanks to
neural encoder-decoder models
\citep{bahdanau2014neural,Sutskever-ea-2014}, Transformer-based
architectures \citep{vaswani2017attention}, and large-scale pretraining
\citep{bertsum,pmlr-v119-zhang20ae,bart},
the past few years have witnessed a huge leap forward in summarization
technology. \textit{Abstractive} methods fluently paraphrase the main
content of the input, using a vocabulary different from the original
document, while \textit{extractive} approaches are less creative ---
they produce summaries by identifying and subsequently concatenating
the most important sentences in a document --- but manage to avoid
hallucinations, false statements and inconsistencies.


Neural extractive summarization is typically formulated as a sequence
labeling problem \citep{cheng2016neural}, assuming access to (binary)
labels indicating whether a document sentence should be in the
summary. In contrast to the plethora of datasets (see
Section~\ref{sec:experiments} for examples) available for abstractive
summarization (typically thousands of document-abstract pairs), there
are no large-scale datasets with \textit{gold} sentence labels for
extractive summarization.  \textit{Oracle} labels are thus
extrapolated from abstracts via heuristics, amongst which greedy
search \citep{nallapati2017summarunner} is the most popular by far
\citep{bertsum,xu2020discourse,dou2020gsum,jia2022neural}.

In this work we challenge received wisdom and rethink whether  greedy search is the best way to  create sentence labels for extractive summarization. Specifically, we highlight two flaws with greedy labeling: 
(1) the search procedure is \textit{suboptimal}, i.e.,~it does not guarantee a global optimum for the  search objective, 
and (2) greedy oracles are \textit{deterministic}, i.e.,~they yield  a \textit{single} reference extract for any given input by associating sentences in the document to its corresponding abstract.

Perhaps an obvious solution to the suboptimality problem would be to
look for oracle summaries following a procedure based on beam search.
Although beam search finds better oracles, we empirically observe that
summarization models trained on these do not consistently improve over
greedy oracles, possibly due to the higher risk of under-fitting
\citep{refresh} --- there are too few positive labels. Moreover, beam
search would also create deterministic oracles.  A summarization
system trained on either greedy or beam oracles is optimized by
maximizing the likelihood of a \textit{single} oracle summary. This
ignores the fact that there can be multiple valid summaries for an
article, in other words, the summary hypothesis space is a naturally
multi-modal probability distribution. We illustrate this point
Table~\ref{tab:example}.


In this paper
we define a new learning objective for extractive summarization which promotes non-deterministic learning in the summary hypothesis space, and introduce \textsc{Oreo}, \textsc{OR}acle \textsc{E}xpectati\textsc{O}n labeling, as a simple yet effective sentence labeling scheme. 
We prove the equivalence between estimating \textsc{Oreo} labels and  optimizing the proposed  learning objective. As a result,  it is sufficient for 
current models to be trained on \textsc{Oreo} labels without requiring any architectural changes. 

Extensive experiments on summarization benchmarks show that
\textsc{Oreo} outperforms comparison labeling schemes in both
supervised and zero-shot settings, including cross-domain and
cross-lingual tasks.  Additionally, we showcase that extracts created
by \textsc{Oreo} can better guide the learning and inference of a
generative system, facilitating the generation of higher-quality
abstracts. We further analyze \textsc{Oreo}'s behavior by measuring
\emph{attainable summary knowledge} at inference time, and demonstrate it is
superior to related deterministic and soft labeling schemes, which we
argue contributes to its consistent performance gain across
summarization tasks.

\begin{table*}[t]
\centering
\footnotesize

\caption{\label{tab:example} 
Sentence labels for a CNN/DM article according to different labeling schemes. Only the first 10 document sentences are shown. 
{Greedy} and {Beam} create oracle summaries (i.e.,~sentences with
label~1) with greedy and beam search, respectively. \textsc{Oreo}, our
labeling algorithm, 
incorporates information from multiple  summary hypotheses shown in the bar chart ($\mathcal{R}$ is the mean of \textsc{Rouge-1} and \textsc{\textsc{Rouge-2}}). 
\textsc{Oreo} assigns high scores to
sentences~2 and~4 which contain an important named entity, \uwave{Jasmine Coleman}, 
and location, \underline{Croydon, South East London}.
In comparison, greedy and beam labeling
consider only one oracle summary, and assign zero to sentences 2 or 4, failing to 
capture that these are informative and should be probably included in the summary. 
}
\begin{scriptsize}
\begin{tabular}{p{0.1cm}p{9cm}p{0.7cm}
p{0.6cm}
p{0.6cm}}\\
\thickhline
\textbf{ID} & \textbf{Document Sentence} & 
\textbf{Greedy} 
& \textbf{Beam} 
&\textbf{\textsc{Oreo}}\\
\hline
1 & 
\uwave{Jasmine Coleman}, 12, has been found safe and well some 50 miles from her home.
& 0 
& 1 
& 0.568 \\
2 & A 12-year-old girl who went missing from her family home at 2 AM amid fears she was driven away by an ``older man'' has been found safe and well. & 1 
& 0 
& 1.000 \\
3 & Jasmine Coleman was reported as missing this morning after disappearing from her home in lancing, west Sussex. & 0 
& 0  
& 0.429 \\
4 & The child was found this afternoon following a police appeal some 50miles away in \underline{Croydon, South East London}. & 1
& 0 
& 0.778 \\
5 & 
Police feared she may have been driven to London by an older man when they launched an appeal for information this morning.
& 1
& 1 
& 0.459 \\
6 & The schoolgirl had not been seen since 11:30 PM on Friday night. & 0 
& 0 
& 0.000 \\
7 & Sussex police said she may have been talking with 
someone on Facetime before disappearing at around 2 am. & 0
& 0 
& 0.555 \\
8 & The force launched a public appeal for information on her whereabouts on Saturday morning. & 0 
& 0 
& 0.000 \\
9 & In it, she was described as fair with long, blonde hair and as having possibly been wearing black riding trousers and a polo shirt or a paisley pattern dress. & 0 
& 0 
& 0.000 \\
10 & On Saturday afternoon the force confirmed she had been found safe and well in Croydon but could not confirm the circumstances under which police located her. & 0 
& 0 
& 0.000 \\
\thickhline
\end{tabular}
\end{scriptsize}
\begin{tabular}{m{6cm}c}\\
\begin{small}
  \begin{center}
    \textbf{Reference Summary}\end{center}
\begin{itemize}[leftmargin=*]
    \item \uwave{Jasmine Coleman} disappeared from her home at around 2 AM this morning.
    \item Police believed she may have been driven towards London by an older man.
    \item She has been found safe and well in \underline{Croydon, South East London} today.
\end{itemize}
\end{small}
& 
\begin{minipage}{0.8\textwidth}
  \pgfplotstableread[col sep=semicolon]{
Rank;Ratio
1 5;57.27699481
2 4 5;51.80351163
1 4 7;50.48192722
2 4 7;50.47348438
1 2 4;50.28058313
2 5;50.01624382
2 3 7;49.73717099
2 3 4;49.2900852
2 4;49.07894687
1 2 7;47.9352911
1 7;47.38927689
2 7;46.98734128
2 3;46.3963959
3 4 5;46.02630809
1 4;45.60275496
3 5 10;45.22336723
}\datatable

\begin{tikzpicture}[inner frame sep=0]
\begin{axis}[
    ybar,
    bar width=4pt,
    width=0.5\columnwidth,
    height=0.7in,
    scale only axis,
    ylabel = $\mathcal{R}$,
    xlabel = {Top-16 Beams (Beam Size $k=256)$},
    ymin=0, ymax=60,
    symbolic x coords={1 5, 2 4 5, 1 4 7,2 4 7,1 2 4,2 5,2 3 7,2 3 4,2 4,1 2 7,1 7,2 7,2 3,3 4 5,1 4,3 5 10},
    xtick={1 5, 2 4 5, 1 4 7, 2 4 7, 1 2 4, 2 5, 2 3 7, 2 3 4, 2 4, 1 2 7, 1 7, 2 7, 2 3, 3 4 5, 1 4, 3 5 10},
    x tick label style={font=\footnotesize,rotate=90,anchor=east},
    legend pos=north east,
    ymajorgrids=true,
    grid style=dashed,
    style={font=\small}
]

\addplot[
    draw=red,
    pattern=north east lines, 
    pattern color=red
    ]
    table[x=Rank, y=Ratio] {\datatable};

\end{axis}
\end{tikzpicture}
\vspace{-1em} 

\end{minipage}
\end{tabular}
\end{table*}

\section{Related Work}


\citet{refresh} were among the first to discuss problematic aspects of
sentence labeling schemes for extractive summarization.  They argue
that labeling sentences individually as in \citet{cheng2016neural}
often generates too many positive labels which leads to overfitting,
while a model trained on greedy labels
\citep{nallapati2017summarunner} underfits the data.
Although extractive performance can be boosted via finetuning pretrained encoders with greedy labels \citep{bertsum},
\citet{matchsum} show that reranking summary candidates constructed
from greedy predictions
can further improve summary quality.
This  demonstrates that the underfitting problem caused by greedy labels still exists even when pretrained models are used. 
Issues with greedy labeling  have also been  discussed from the perspective of  \textit{data bias},
including {lead bias} \citep{ani2005lead,kedzie2018content,grenander2019countering}  --- greedy labels display a bias towards lead sentences in news text and systems trained on them do not easily transfer to other domains --- 
and {monolingual bias} \citep{jia2022neural} ---
 greedy labels created for one language (e.g., English) do not transfer to a different language.




The idea of learning from multiple references has found application in
various tasks including dialog response generation
\citep{gao2019jointly}, machine translation
\citep{khayrallah2020simulated}, and question answering
\citep{zhu2020question}. Summarization datasets with multiple
references are not generally available for model training, but a few have been manually created for system evaluation  \citep{dang2005overview}. 
In extractive summarization, gold references in the form of sentence
labels do not usually exist, and learning from multiple references has
not been yet explored.  In this work, we use beam search to
create multiple high-quality \textit{oracle} summaries, from which
summary-level supervision is aggregated into sentence labels to
promote multi-reference learning for extractive summarization.

\section{Problem Formulation}
\label{sec:formulation}
Let $D=\{x_i\}_1^m$ denote a document consisting of sentences $x_i$.
An extractive summarizer produces a \textit{summary hypothesis} that
represents salient information via identifying a subset of sentences
$\hat{Y}=\{\hat{y}_j\}_1^n, n<<m$ within~$D$.  In practice, ROUGE
\citep{lin2003automatic}, an automatic metric based on lexical overlap
is commonly adopted to evaluate~$\mathcal{R}(\hat{Y}, S)$, the quality
of~$\hat{Y}$ against gold \textit{summary reference}~$S$.

Following previous work \citep{cheng2016neural,nallapati2017summarunner,refresh,bertsum}, we conceptualize extractive summarization as a sequence labeling task, and aim to build a system that estimates the summary-worthiness of each sentence in a non-autoregressive manner. 
As mentioned earlier, sentence labels need to be first extrapolated to train an extractive system, since
existing datasets are label-free, they only contain document-abstract pairs.  
\textsc{BertSum} \citep{bertsum} is a popular extractive model and representative of the approach sketched above.  
Built on top of BERT \citep{devlin2019bert}, it adds a two-layer
Transformer \citep{vaswani2017attention} for sentence representation
and a \texttt{sigmoid} layer for summary prediction.  During
inference, document sentences~$\{x_i\}_1^m$ are ranked based on their
estimated scores, and summary~$\{\hat{y}_j\}_1^n$ is identified. The
number of sentences~$n$ to be included in the summary is often
pre-defined and fixed.

\section{From Existing Labeling Schemes to Oreo}
\label{sec:prev_work}
Early labeling methods create sentence labels~$\ell_i$ by evaluating
the similarity of~$x_i$ against reference summary~$S$ through various
heuristics $h(\cdot)$, $\ell_i \overset{\mathrm{def}}{=} h(x_i, S)$,
including \textsc{Rouge} \citep{lin2003automatic} and rule-based
features such as sentence and paragraph position information and the
number of mentioned entities
\citep{woodsend2010automatic,cheng2016neural}.
These methods obtain \textit{local} labels as they assume a sentence
can be classified as summary-worthy based on its own, without taking
the summary context into account.

However, model evaluation does not operate at the sentence-level, 
as a sentence has to form a summary hypothesis~${Y}$ together with other
candidates \citep{refresh}. The aim of extractive summarization is to
deliver a high-quality summary hypothesis, i.e.,~a good set of
sentences rather than a set of good sentences.  A sentence might
achieve a high score on its own but contribute little to a summary
hypothesis, e.g., due to redundancy.
An alternative is to obtain \textit{global} labels, based on whether a
sentence occurs within the optimal set of sentences which collectively
achieve the highest score according to some evaluation metric like
\textsc{Rouge} \citep{lin2003automatic}:
\begin{equation}
    \ell_i \overset{\mathrm{def}}{=} \mathbbm{1}(x_i \in {Y}^{\ast})
    \text{ where }
    {Y}^{\ast} = \argmax_{Y\in \BBC(D)} \mathcal{R}(Y,S).
    \label{eq:best}
\end{equation}
where $|\BBC(D)|=C{m \choose n}$ is the hypothesis combinatorial
space.  As Equation~(\ref{eq:best}) is computationally intractable, in
practice, it is approximated by further conditioning on a heuristic
search space~$\BBS$ such that~${Y}^{\ast} \approx \argmax_{Y \in
  \BBS(D)} \mathcal{R}(Y,S)$,
and the approximated $Y^\ast$ is usually called \textit{oracle}
summary.  A widely adopted approximation is \textit{greedy} labeling
\citep{nallapati2017summarunner,refresh,bertsum,matchsum}, which uses
greedy search to maximize $\mathcal{R}$ at each step of sentence
selection (the algorithm stops when~$\mathcal{R}$ can no longer
increase or the maximum number of steps is reached).  We present the
greedy labeling algorithm in Appendix~\ref{appendix:greedy}.

While significantly reducing  complexity,
greedy labeling does not guarantee a global optimum.  To find better
summaries to serve as oracles, we propose to replace greedy search
with beam search which we refer to as \textit{beam} labeling.  We
empirically find that around $8$\%--$20\%$ of (greedy) labels can be
potentially improved with beam search (when setting the beam size
to~256; see Appendix~\ref{appendx:beam} for details).
However, having better labels does not necessarily translate to performance improvements, and we  discuss why this is the case next.

\subsection{\textsc{Oreo}: Estimating Oracle Expectation}
\label{sec:oreo}
Extractive summarization models are typically trained to optimize
$\max p_\theta (Y^\ast|D)$, where the best hypothesis $Y^\ast$ can be
approximated with greedy or beam search.  This learning objective
maximizes the probability at a {single} point $Y^\ast$, and assigns
zero probability to other summary hypotheses $\hat{Y}$, regardless of
their quality. We note that this formulation leads to a
\textit{discrepancy} between how the model is optimized and how the
labels against which this optimization takes place are obtained.
Given an input document,
sequence labeling summarization models assume conditional independence 
at sentence-level inference, while in greedy labeling, each step in
the process of maximizing $\mathcal{R}(Y^\ast, S)$ conditions on the
outcomes of previous steps.  From an optimization perspective, this
mismatch renders fitting a non-autoregressive sequence labeler
difficult for two reasons: (1) learning to search and maximizing the
likelihood at $Y^\ast$ is challenging, and so the model tends to
underfit~$Y^\ast$ \citep{refresh}, and (2) probabilities at
\textit{other} ${Y}$ with high evaluation scores remain
\textit{under-estimated} and \textit{uncalibrated} due to supervision
sparsity.  Simply replacing greedy search with beam search does not
resolve these optimization challenges as point estimation is still
performed.

A solution is to evaluate summary hypotheses during training and
reward the model accordingly \citep{refresh}.  However, this is
non-trivial as the the metric $\mathcal{R}$ is usually
non-differentiable, and it is computational expensive to sample from a
large combinatorial hypothesis space, e.g.,~with Reinforcement
Learning \citep{sutton2018reinforcement}.
Rather than changing the training of the model, in this work, we study
how to derive a better sentence labeling algorithm that leads to a
better optimization objective.

Specifically, we wish to incorporate multiple high-quality hypotheses as oracle summaries into the learning objective. Our key assumption is that extractive oracles are non-deterministic, but  drawn from a distribution $p({Y}^\ast | D, S)$. We thus formulate the objective for extractive summarization as:
\begin{gather}
    \max \EXP_{{Y}^\ast \sim p({Y}^\ast|D, S)} 
    \left[ 
    \mathcal{R}(Y^\ast, S)
    p_\theta (Y^\ast|D) 
    \right].
    \label{eq:obj}
\end{gather}

Under this formulation, an optimized model is expected to assign high probability $p_\theta({Y} | D)$ when there exists an oracle summary with high probability 
and high score according to some quality evaluation metric.


From the perspective of sentence labeling,
we note that a candidate~$x_i$ relates to the summarization task  through the oracle summary space~$\Y$. As~$\Y$ is a combinatorial space, the mapping $x_i \rightarrow Y^\ast$ is one-to-many. 
Therefore, we can compute the probability for each candidate to be selected via marginalization: 
\begin{equation}
\label{eq:margin}
    p(x_i | D, S) 
    = \sum_{Y^\ast}^\Y
    p(x_i, Y^\ast|D, S)
    = \sum_{Y^\ast}^\Y
    {p(x_i | Y^\ast, D)}
    {p(Y^\ast|D, S)}.
\end{equation}

To connect marginalization in Equation~(\ref{eq:margin}) with the
summarization objective in Equation (\ref{eq:obj}), we further
incorporate hypothesis evaluation $\mathcal{R}(Y^\ast, S)$, and define
the summary-worthiness of a sentence~$x_i$ as the expectation of its
associated oracle evaluation:
\begin{equation}
    \ell'_i
    \overset{\mathrm{def}}{=} \sum_{Y^\ast}^\Y
    {\mathcal{R}(Y^\ast, S)}
    {p(x_i | Y^\ast, D)}
    {p(Y^\ast|D, S)}
    = \EXP_{
    Y^\ast \sim 
    \underbrace{\scriptstyle p(Y^\ast|D, S)}_{\text{oracle distribution}}}
    \left[ 
    \underbrace{\mathcal{R}(Y^\ast, S) }_{\text{oracle evaluation}}
    \underbrace{p(x_i | Y^\ast, D)}_{\text{oracle membership}}
    \right]\label{eq:v}
\end{equation}
where the oracle membership $p(x_i | Y^\ast, D)$ is identical to $y_i
= \mathbbm{1}(x_i \in {Y}^\ast)$ and the oracle distribution will be
discussed in Section~\ref{sec:oracle_dist}.  Given a sequence labeling
model~$\theta$, maximizing the oracle expectation for all input
sentences is equivalent to the objective in
Equation~(\ref{eq:obj}). We present the proof in
Appendix~\ref{appendix:proof}.

 To be compatible with the standard sequence labeling architecture for extractive summarization, we perform MLE with a cross-entropy loss:
\begin{equation}
    \min \mathcal{L}(\theta) = \min
    \sum_{i=1}^m
    \mathrm{CrossEntropy}\left(\ell(x_i), p_\theta(x_i|D,S)\right) \label{eq:final_obj}
\end{equation}
where the scaled expectation $ {\ell}(x_i) =
({\ell}'_i-\bar{\ell}_{\text{min}})/(
\bar{\ell}_{\text{max}}-\bar{\ell}_{\text{min}})$ constitutes the
final sentence labels.  The details of oracle expectation labeling are
given in Algorithm~\ref{algo}.

\algtext*{EndWhile}
\algtext*{EndIf}
\algtext*{EndFor}


\begin{algorithm}[t]
\small
\caption{\label{algo}
Labeling with Oracle Expectation}

\vspace{-1.5em}
\renewcommand{\algorithmiccomment}[1]{{\color{gred}
\hfill$\triangleright${ #1}}}
\begin{multicols}{2}

\begin{algorithmic}[1]
\Function{Oreo}{$n$, $k$, $p$}
\newline
\Comment{Max number of sentences in a summary,
beam size, and oracle distribution}
\State Initialize beam $\mathcal B$
\For{$j \gets n$}
    \State $\mathcal B \gets$  \Call{Step}{$j$, $\mathcal B$}
\EndFor

\State Initialize $\ell'_i$ to $0$, $\forall i$  
\Comment{Pre-scaled  expectation}
\For{$b, r \gets \mathcal{B}$}
\For{$x \gets b$}
    \State $\ell'_i \gets r + \ell' * p_i$
\EndFor
\EndFor
\State $\ell =$ \Call{MaxMinScale}{$\ell'$}
\State \Return $\ell$
\EndFunction
\end{algorithmic}

\columnbreak

\begin{algorithmic}[1]
\Function{Step}{$j$, $\mathcal B$}
\Comment{Step and beam}
\State Initialize visited paths $\mathcal V$
\For{$b, v \gets \mathcal{B}$}
    \If{$|b|<j$}  
    \State \algorithmiccontinue \Comment{Skip early stopping}
    \EndIf
    \For{$i \gets |D|$}
        \State $b' =$ \Call{Sort}{$b + \{i\}$}
        \If{$b'$ not in $\mathcal V$}
        \State $v' =$ \Call{Rouge}{$b'$}
            \If{$v' > v$}
            \State $\mathcal B \gets \mathcal B + \{(b', v')\}$
            \EndIf
        \EndIf
        \State $\mathcal V \gets \mathcal V + \{b'\}$
    \EndFor
\EndFor
\State \Return \Call{Top-$k$}{$\mathcal B$} \Comment{Pruned beam}
\EndFunction
\end{algorithmic}

\end{multicols}

\vspace{-1em}

\end{algorithm}



\subsection{Comparison with Existing Labeling Algorithms}
\textsc{Oreo} creates soft (continuous) sentence labels, i.e.,~it
incorporates summary-level evaluation while maintaining low sparsity.
A detailed comparison with other labeling algorithms is provided in
Table~\ref{tab:algo}.
Equation~(\ref{eq:v}) also bears a resemblance to the RL objective used in \citet{refresh}: 
$\max \EXP_{\hat{Y} \sim p_\theta({Y}|D)}
[\mathcal{R}(\hat{Y},S)]
$.  \citet{refresh} evaluate summary hypotheses directly while
Equation~(\ref{eq:v}) estimates sentence-level membership.  This is a
consequence of the nature of the sequence labeler which does not
explicitly represent the summary hypothesis space (which is
combinatorial), and therefore supervision is delegated to sentences
rather than summaries.  By maximizing sentence-level likelihood,
estimations for the associated hypotheses are updated, albeit
indirectly.

\citet{refresh} employ REINFORCE \citep{williams1992simple}, an
on-policy RL algorithm that samples from a model during training,
while in Equation~(\ref{eq:v}), samples are drawn from the
non-parametrized oracle distribution $p(Y^\ast|D,S)$.  We provide
\textit{offline} supervision in the form of static sample labels. In
contrast to the \emph{online} reward in RL, offline labels can be
reused during the entire course of training and are therefore more
sample efficient.  Our labeling scheme can be seen as a type of
offline bandit learning \citep{tang2022offline}.
    While offline RL has been recently applied to abstractive summarization \citep{pang2021text}, it  remains under-explored in extractive summarization.
    


\begin{table}[t]
\centering
\small
\setlength{\tabcolsep}{4pt}
\def\arraystretch{1.2}


\caption{\label{tab:data_stats}
  Datasets for monolingual and cross-lingual (last column) summarization. Compression rate denotes the number of sentences extracted to form a summary; and $\dagger$ denotes that  trigram blocking \citep{bertsum}
  was applied in sentence selection for redundancy removal.}
  \begin{tabular}{lrrrrrr}
\thickhline
Datasets & CNN/DM & XSum & Multi-News  & Reddit & WikiHow & MLSum\\  \thickhline
Language & En & En & En & En & En & En/De/Es/Fr/Ru/Tr\\
Domain & Newswire & Newswire & Newswire & Social Media & Wikipedia
& Newswire\\
\#Train &287,084  & 203,028 & 44,972 & 41,675 & 168,126 & 287,227 (En)\\
\#Validation &13,367  & 11,273 & 5,622 & 645 & 6,000 & 13,368 (En)\\
\#Test &11,489  & 11,332 & 5,622 & 645 & 6,000 & 53,981 (Non-En)\\
\#Compression Rate & 3$^\dagger$  & 2 & 9 & 2 & 4$^\dagger$ & 2$^\dagger$\\
\thickhline
\end{tabular}
\end{table}

\subsection{The Oracle Distribution}
\label{sec:oracle_dist}

We derive the oracle distribution $p(Y^\ast|D)$ heuristically bearing
in mind that: (a)~we have no prior knowledge as to which hypothesis is
more or less likely as an oracle summary and therefore assume the
oracle distribution to be uniform over a large hypothesis space; and
(b)~it is desirable for~$p(Y^\ast|D)$ to positively correlate with $\mathcal R(Y^\ast, S)$ and we expect
this correlation to become stronger over the course of optimization. 
In practice, we use beam search (with beam size $k<<|\Y|$) to find potential oracle summaries, and adopt a uniform distribution over top-ranked beams:
$p(Y^\ast|D) \sim {U}(1, t)$, 
where $t<k$ is a hyper-parameter which we optimize on a development set.
To further incorporate 
other oracle features, we also experimented with several weight
annealing mechanisms over top beams as determined by~$\mathcal{R}$, our
summary quality evaluation metric (see Appendix
\ref{appendix:oracle_dist} for details).

\section{Experiments}
\label{sec:experiments}
\subsection{Supervised Extractive Summarization}

We conducted all extractive summarization experiments with
\textsc{BertSum} \citep{bertsum}, the neural summarization
architecture introduced in Section~\ref{sec:formulation}.  We opted
for \textsc{BertSum}  due to its simplicity and
popularity in a wide range of summarization tasks
\citep{matchsum,xu2021generating,jia2022neural}.  We nevertheless note
that \textsc{Oreo} is \textit{model-agnostic} and can  be
also applied to more complex architectures.
For a fair comparison between different labeling schemes, we follow
the standard training configuration used in \citet{bertsum} without
any additional hyper-parameter optimization (e.g., for our specific
labeling scheme).  We set~$\mathcal{R}$, the summary quality
evaluation metric, to the mean of \textsc{Rouge-1} and
\textsc{Rouge-2}. We report experiments on a variety of summarization
datasets including CNN/DM \citep{hermann2015teaching}, 
XSum \citep{xsum}, 
Multi-News
\citep{fabbri2019multi},
Reddit \citep{reddit}, and
WikiHow \citep{koupaee2018wikihow}. Detailed statistics are shown in
Table~\ref{tab:data_stats}.


\begin{table}[t]
  \small
  \begin{tabular}{l@{\hspace{.5cm}}l}
    \centering
 \begin{minipage}[t]{2.4in}
   \caption{\label{tab:algo} Sentence labeling schemes for extractive
     summarization. {Sum} refers to summary-level evaluation. $m$, $n$,
     and $k$ respectively denote document size, summary size, and beam
     size.}
\begin{tabular}{@{}l@{~}c@{~}l@{~}l@{}}
\thickhline
{Scheme}
& {Sum}
& {Sparsity} 
& {Complexity}\\
\hline
Local
& \xmark
& Medium  
& $\mathcal{O}(m)$
\\
Global 
& \cmark 
& High  
& $\mathcal{O}(
\frac{m!}{n! (m - n)!})$\\
Greedy 
& \cmark 
& High   
& $\mathcal{O}(n m \log m)$\\
\hline
Beam (ours) 
& \cmark 
& High 
& 
$\mathcal{O}(nmk \log (mk))$\\
\textsc{Oreo} (ours) 
& \cmark 
& Low 
& $\mathcal{O}(nmk \log (mk))$\\
\thickhline
\end{tabular}

 \end{minipage} &
 \begin{minipage}[t]{2.7in}
   \centering

\caption{\label{tab:results_mono} Extractive performance (test set,
  \textsc{Rouge-L}) on {CNN/DM} ({CM}), XSum ({XS}), Multi-News
  ({MN}), Reddit ({RD}), and WikiHow ({WH}).  We highlight
  \textbf{highest} and \textit{lowest} scores.  }
\begin{tabular}{@{}l@{~~}c@{~~}c@{~~}c@{~~}c@{~~}c@{}}
\thickhline
{Systems}& 
{CD}& 
{XS}&
{MN}&
{RD}&
{WH}
\\
\hline
{\sc Lead} 
&36.67  
&14.79
&38.97 
&14.34
&23.24\\
{\sc MatchSum}
&40.38
&18.41
&41.89
&20.13
&29.58\\
\hline
{\sc Oracle} \\
~~~~Greedy
&48.87 
&23.57 
&44.27 
&28.98
&32.68
\\
~~~~Beam
&52.86 &23.71 &46.40
&29.11 &36.51
\\
~~~~\textsc{Oreo}
&50.08 &20.07 &46.14 
&24.55 &34.28
\\
\hline
{\sc BertSum}\\
~~~~Greedy
&\textit{39.56} &
\textit{17.16} &
41.53 &
\textit{19.11} & 
{28.24}
\\
~~~~Beam
&{39.66}
&{17.66}
&\textit{41.50}
&19.81
&\textit{25.71}\\
~~~~\textsc{Oreo}
&\textbf{39.96}
&\textbf{17.81}
&\textbf{41.71}
&\textbf{20.02}
&\textbf{28.46}
\\
\thickhline
\end{tabular}

    \end{minipage} 
  \end{tabular}
   \vspace{-1.4cm}

\end{table}


Our results are presented in Table~\ref{tab:results_mono}.  In the
first block, we report the performance of the \textsc{Lead} baseline
which considers the first~$k$ sentences in a document as the summary
(see last row in Table~\ref{tab:results_mono}) and \textsc{MatchSum}
\citep{matchsum}, a state-of-the-art system which performs summary
reranking with another BERT model.  The second block reports
\textsc{Oracle} performance with greedy labels, beam labels ($k=256$),
and \textsc{Oreo} labels ($k=256, t=16$; we take the top-$n$ sentences
with non-zero scores). 
 See Appendix \ref{appendx:setting} for the
labeling hyperparameters $k, t$ for each dataset, and more detail on experimental settings. 
The third block reports \textsc{BertSum} performance with
different labeling schemes.

Although beam \textsc{Oracle} is superior to greedy \textsc{Oracle}
and raises the upper bound, the overall performance of
\textsc{BertSum} optimized on beam labels does not significantly
improve upon its greedy counterpart.  In fact, performance drops
drastically on WikiHow.  \textsc{Oreo} shows inferior \textsc{Oracle}
results as it considers multiple top-ranked beams and is therefore not
bound-preserving (see Section~\ref{sec:analysis} for detailed
analysis).  However, \textsc{BertSum} trained with \textsc{Oreo}
labels consistently outperforms a model trained with beam labels, and
achieves a~0.18--0.89 \textsc{Rouge-L} improvement on different
benchmarks compared to greedy labeling.  Although \textsc{BertSum}
with \textsc{Oreo} still falls short of the state-of-the-art, we show
that \emph{one-stage} summarization modeling can be enhanced with
better labeling, which can potentially serve as a foundation for more
complex reranking methods.

\subsection{Zero-Shot Cross-Domain Extractive Summarization}
\input{fig_lead_bias_small}


We further examine the generalization capability of models trained
with \textsc{Oreo} labels in a \textit{zero-shot}
setting. Specifically, we evaluate a model trained on CNN/DM, against
XSum, another news summarization dataset with shorter summaries (at
most 2 sentences), and Reddit and WikiHow which represent entirely
different domains (discussion forums and instructional text) and
topics.

Table~\ref{tab:results_xdom} summarizes our results.  Models trained
with \textsc{Oreo} perform on par with greedy labeling in-domain but
display stronger generalization cross-domain. Greedy labels are more
prone to lead bias, they deem as summary-worthy sentences mostly from
the beginning of the document. Such bias is present in news datasets
like CNN/DM but does not transfer to other domains like social media
or Wikipedia. \textsc{Oreo} alleviates this bias and performs better
out-of-domain. As shown in Figures~\ref{fig:lead_bias}(a)
and~\ref{fig:lead_bias}(b), \textsc{Oreo} is less concentrated on lead
sentences in news text.



\begin{table}[t]
  \small
  \begin{tabular}{l@{~~}l@{\hspace{.5cm}}l@{}}
\begin{minipage}[t]{1.5in}
\centering
     \caption{\label{tab:results_xdom}
 Cross-domain performance  for models trained on
 {CNN/DM} (zero-shot; \textsc{Rouge-L}). We highlight \textbf{highest} and \textit{lowest} performance.}

\begin{tabular}{@{}l@{~}c@{~}c@{~}c@{}}
\thickhline
{CNN/DM} & 
~~{XS}&
~~{RD}&
~~{WH}\\
\hline
{\sc BertSum}\\
~~~~Greedy
&\textbf{15.62} &\textit{17.06} 
&25.39\\
~~~~Beam
&\textbf{15.62}
&17.64
&\textit{24.77} \\
~~~~\textsc{Oreo}
&\textit{15.58}
&\textbf{17.71}
&\textbf{25.62}
\\
\thickhline
\end{tabular}


  \end{minipage}
&
\hspace*{.5ex}\begin{minipage}[t]{2.4in}
\centering
\caption{\label{tab:result_xl} Zero-shot cross-lingual summarization
  on {MLSum} (test set, \textsc{Rouge-L}). Systems with~$^\ast$
  are supervised. Systems with~$^\dagger$ use XLM-R large. 
}

\begin{tabular}{@{}l@{~}c@{~}c@{~}c@{~}c@{~}c@{~}c@{}}
\thickhline
{{Systems}} 
& {De} & {Es} & {Fr} & {Ru} & {Tr} & {AVG}
\\
\hline
{\sc XlS$^{\ast\dagger}$}
&41.28
&21.99
&24.12
&10.44
&33.29
&26.22\\
\textsc{NLS}$^\dagger$
& 34.95
&21.20
&23.59
&10.13
&31.49
&24.27\\
\hline
{\sc XlS}\\
~~~~Greedy
&28.75	
&\textit{20.83}	
&{23.10}
&9.43
&\textit{29.52}	
&22.33
\\
~~~~{Beam}
&\textit{26.43}
&\textbf{20.90}
&\textbf{23.41}
&\textit{9.42}	
&29.80	
&\textit{21.99}
\\
~~~~\textsc{Oreo}
&\textbf{31.47}
&20.84	
&{23.10}	
&\textbf{9.44}
&\textbf{31.71}
&\textbf{23.31}\\
\thickhline
\end{tabular}


\end{minipage}
&
\hspace*{-2ex}\begin{minipage}[t]{1.4in}
\centering
\caption{\label{tab:results_abs}
Results for abstractive summarization  on {CNN/DM} (test set). 
 \textsc{R}-1/2/\textsc{L} is a shorthand for \textsc{Rouge}. 
}

\begin{tabular}{@{}l@{~}c@{~}c@{~}c@{}}
\thickhline
{Systems}
& \textsc{R-1} & \textsc{R-2} & \textsc{R-L}\\
\hline
\textsc{Bart}
& 44.16 & 21.28 & 40.90 \\
\hline
\textsc{GSum}
\\
~~~~Greedy
& \textit{44.40} & \textit{21.52} & \textit{41.23} \\
~~~~Beam
& 44.41 & 21.55 & 41.26 \\
~~~~\textsc{Oreo}
&\textbf{44.81}
&\textbf{21.83}
&\textbf{41.60}
\\
\thickhline
\end{tabular}

\end{minipage}
  \end{tabular}
 \end{table}

\subsection{Zero-Shot Cross-Lingual Extractive Summarization}
We next investigate the generalization capabilities of our approach
in a cross-lingual setting.  We use English data for model training
and report \textit{zero-shot} results on a variety of languages from
the MLSum dataset (\citealt{mlsum}; see Table~\ref{tab:data_stats} for
detailed statistics).  Following \citet{jia2022neural}, we augment
English articles with word replacement during training
\citep{qin2021cosda} using the MUSE \citep{lample2018word} bilingual
dictionary to align multilingual representations. We adopt a word
replacement rate of~0.5.  \textsc{BertSum} was initialized with XLM-R
base \citep{xlmr}, a cross-lingual pretrained model (see \textsc{XlS}
in Table~\ref{tab:result_xl}).\footnote{We also
experimented with mBERT \citep{devlin2019bert} and achieved similar
results. See Appendix~\ref{appendix:more_results}.}

The first block in Table~\ref{tab:result_xl}, reports the results of a
\textit{supervised} \textsc{XlS} model which has access to training
data for \textit{all} languages; \textsc{NlS} is the zero-shot state
of the art system of \citet{jia2022neural}; their approach creates
multiple sets of greedy labels with different machine translation
methods and adopts a neural architecture to learn weights for the
obtained label sets. The second block presents the results of a
zero-shot \textsc{XlS} model with different labeling schemes. As can
be seen, \textsc{Oreo} labels on Spanish, French, and Russian are on
par with greedy labeling.  Systems trained with greedy labels exhibit
less cross-lingual generalization on German and Turkish, while
\textsc{Oreo} improves system performance on German by 2.72
\textsc{Rouge-L} points and on Turkish by~2.19.  Previous work
\citep{jia2022neural} shows that cross-lingual performance correlates
with lead bias in the target language. For example, Turkish articles
are less lead-biased than Russian in MLSum, and thus benefit more from
better sentence labeling.  \textsc{Oreo} trails behind \textsc{NlS}
which is not surprising as the latter model benefits from more
resources, i.e.,~machine translation and XLM-R large, and a more
complex network architecture.

\subsection{Supervised Abstractive Summarization}

We further assessed whether the proposed labeling scheme is of benefit
to abstractive summarization. We experimented with \textsc{GSum}
\citep{dou2020gsum}, a state-of-the-art abstractive system that takes
extractive summaries as additional input to \textit{guide} the
generation of document abstracts.  During training, \textsc{GSum} uses
extractive oracles as guidance, while at inference time guidance is
provided by summary hypotheses produced by a trained extractive
system.  We initialized \textsc{GSum} with BART \citep{bart}, and used
\textsc{BertSum} as the guidance model optimized with different
labeling schemes (i.e.,~greedy, beam and \textsc{Oreo}).


Abstractive summarization results are shown in
Table~\ref{tab:results_abs}.
The first block shows the performance of \textsc{Bart} \citep{bart}
which serves as a baseline.  In the second block, we report the
performance of \textsc{GSum} \citep{dou2020gsum} with greedy labels (default) in addition to beam- and \textsc{Oreo}-based
variants.  As we can see, while beam labeling performs on par with its
greedy counterpart, \textsc{Oreo} guidance boosts performance with
0.37 ROUGE-L points over vanilla \textsc{GSum}.  We conclude that
abstractive systems can also benefit from our expectation-based
labeling algorithm, without any modeling changes or hyperparameter
optimization.  More results with varied guidance settings can be found
in Appendix~\ref{appendix:more_results}. Examples of system output are
shown in Appendix~\ref{appendix:sys_out}.


\subsection{Comparison with Bound-Preserving Methods}
\label{sec:analysis}

Let $\{z^\ast_i\}_1^m, z^\ast_i \in \{0, 1\}$ denote a multi-hot
representation of an oracle summary.  We define a labeling function
$\ell$ as \textit{bound-preserving}, if there exists a constant
$\gamma \in [0,1]$ so that the condition
$\mathbbm{1}\left(\ell(x_i)>\gamma \right) = z^\ast_i, \forall i$
holds.
Intuitively, the condition holds if and only if the top-ranked
sentences remain identical.  Bound preservation of soft labels
guarantees that the performance upper bound of a summarization system
trained on soft labels equals that of a system trained on their
original hard labels, e.g., obtained via  greedy and beam search.
For instance, label smoothing \citep{szegedy2016rethinking}, a common technique for training deep neural networks, 
is  bound-preserving. 
In contrast, \textsc{Oreo} is generally \textit{not}
bound-{preserving} for either beam or greedy oracles.\footnote{Under
two special cases our labeling scheme is bound-preserving: (1) with
beam size $k=1$, \textsc{Oreo} is equivalent to greedy labeling and
(2) with top beam size $t=1$, \textsc{Oreo} is equivalent to beam
labeling.}
To further analyse this property of soft labels, we propose
\mbox{\textsc{Ormax}} as a bound-preserving variant, by replacing the
expectation with a $\max$ operator: ${\ell}'_i
\overset{\mathrm{def}}{=}\max_{Y^\ast \sim p(Y^\ast|D, S)} \left[
  p(x_i|Y^\ast) \mathcal{R}(Y^\ast, S) \right]$.  Compared to
\textsc{Oreo}, \textsc{Ormax} incorporates multiple oracles while
additionally preserving the upper bound of beam
labels.\footnote{\textsc{Ormax} is trivially bound-preserving since
 sentences selected by the top-ranked beam receive the
highest score, and the top-ranked beam can  be reconstructed
by the top-ranked sentences.  We illustrate the difference between
\textsc{Oreo} and bound-preserving methods in Figure~\ref{fig:eps} in
Appendix~\ref{appendix:bound_preservation}.  }



\begin{table}[t]
\small
  \begin{tabular}{@{}l@{\hspace{.8cm}}l@{}}
\raisebox{0in}[0pt]{    \begin{minipage}[b]{2.7in}
      \centering
\caption{\label{tab:results_smoothing}
Comparison of \textsc{Oreo} to bound-preserving labeling (CNN/DM test
set).  Results shown for extractive (\textsc{BertSum}) and abstractive
(\textsc{GSum}) summarization. LS refers to Label Smoothing ($\alpha$
optimized between $\{0.1,0.2,0.3\}$).  \textsc{R-}1/2/\textsc{L} is a
shorthand for \textsc{Rouge}.  } 


\begin{tabular}{lccc}
\thickhline
{Systems}
& \textsc{R-1} & \textsc{R-2} & \textsc{R-L}\\
\hline
\multicolumn{4}{l}{\textsc{BertSum}}
\\
~~~~Greedy 
&43.18 & 20.16  & 39.56 \\
~~~~Greedy+LS
&$\uparrow$0.03 & $\downarrow$0.01 & $\uparrow$0.03\\
\hdashline
~~~~Beam 
&43.25 &20.14 &39.66\\
~~~~Beam+LS
&$\uparrow$0.00 & $\downarrow$0.02 & $\uparrow$0.01\\
\hdashline
~~~~\textsc{Oreo}
&{43.58} &{20.43} 
&{39.96} \\
~~~~\textsc{Ormax}
&$\downarrow$0.20 & $\downarrow$0.27 & $\downarrow$0.20\\
\hline
\multicolumn{4}{l}{\textsc{GSum}}
\\
~~~~\textsc{Oreo}
&44.81
&{21.83}
&{41.60}\\
~~~~\textsc{Ormax}
&$\uparrow$0.04 & $\downarrow$0.02 & $\downarrow$0.00\\
\thickhline
\end{tabular}


    \end{minipage}} &
    \begin{minipage}[t]{2.4in}
      \centering
\hspace*{-.6cm}\includegraphics[width=7cm]{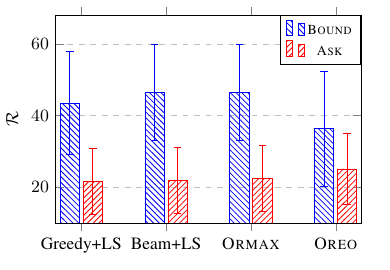}
\captionof{figure}{\label{fig:metric} 
Upper bound and attainable summary
  knowledge captured by sentence labeling method (CNN/DM validation set) for
  Label Smoothing (+LS), \textsc{Ormax}, and \textsc{Oreo}.
}  
    \end{minipage}
  \end{tabular}
  \end{table}

Table~\ref{tab:results_smoothing} shows the performance of label
smoothing and \textsc{Ormax} for extractive (first block) and
abstractive (second block) summarization.
Although label smoothing has been successfully applied to
discriminative \citep{szegedy2016rethinking} and generative NLP tasks
\citep{chen2018best,bart}, the soft labels it creates do not yield
better results than {their original hard labels in extractive
summarization}.
Label smoothing performs \textit{implicit model calibration}
\citep{muller2019does}, which can potentially improve sentence ranking and selection at inference, however, it also imposes regularization in
neural networks training
\citep{szegedy2016rethinking}, which may render it less effective for
extractive summarization where there is a higher risk of underfitting
\citep{refresh}.
On the other hand, \mbox{\textsc{Ormax}} performs on par with
\textsc{Oreo} on abstractive summarization, while
it underperforms on extractive summarization.
Although bound preservation is, intuitively, desirable, 
our experimental results suggest that it is neither a necessary nor
sufficient condition to produce a well-optimized summarization system.


 

\subsection{Performance Analysis}
\label{sec:further}

Previous experiments revealed that oracles are not
necessarily indicative of model performance (see
Tables~\ref{tab:results_mono} and~\ref{tab:results_smoothing}), 
due to the discrepancy between model optimization and sentence
labeling, as discussed in Section~\ref{sec:oreo}.
To further understand how different labeling schemes (and the oracles
based on them) influence model performance, we quantify this
discrepancy via a sampling-based method which simulates  sentence
selection for a  sequence labeling model at inference.

Bearing in mind that a non-autoregressive sequence labeling model
performs conditionally independent predictions and selects a fixed
size number of sentences~$n$, we construct summary hypotheses
$\hat{Y}=\{\hat{y}_j\}_{j=1}^n$ by drawing \textit{independent}
sentence samples, and measure the extent to which a model can attain
summary relevant knowlege (\textsc{ASK}) as:
 \begin{equation}
     \textsc{ASK}
     \overset{\mathrm{def}}{=} \EXP_{\{\hat{y}_j\}_{j=1}^n \sim p(x_i|D,S)}
     \left[ 
     \mathcal{R}(\{\hat{y}_j\}_{j=1}^n, S) \right] 
     \text{ where } 
     p(x_i|D, S) = \frac{\ell_i}{\sum_{i=1}^m \ell_i}
 \end{equation}
 
Note that $p(x_i|D, S)$ is shaped by soft labels, and thus results in varied sentence/summary samples for different labeling schemes.
The comparison in Figure~\ref{fig:metric} explains why we observe
performance gains from \textsc{Oreo} despite obtaining the lowest
upper bound performance.  The latter considers only the best case
scenario at inference, ignoring the fact that some summary knowledge
encoded in sentence labels can be hard or impossible to attain,
e.g.,~when sentences in the oracle summary are highly dependent (and
is therefore challenging to select them jointly with a model making
independent predictions), or the oracle summary contains less than $n$
sentences (which again entails that information is missing).  Compared
to other labeling schemes, \textsc{Oreo} captures richer summary
information that is attainable for sequence labeling models, narrowing
the distance between \textsc{Bound} and \textsc{ASK}.  Consistent
with our analysis, systems trained on \textsc{Oreo} perform robustly
on a wide variety of summarization tasks.

\section{Conclusions and Future Work}
We provided a comprehensive analysis of existing labeling schemes for
extractive summarization, and identified two flaws in greedy labeling,
namely it delivers suboptimal and deterministic labels.
We proposed a novel optimization objective to learn from multiple
oracle summaries, which can be instantiated by a labeling
scheme based on oracle expectation.  Experimental results show that
the proposed  scheme achieves substantial improvement across
domains and languages, without any architectural modifications. 
Our framework is agnostic to the labeling
metric~$\mathcal{R}$, however, an important future direction is to incorporate
different learning signals and provide sentence labels
with more desirable properties, such as query relevance
\citep{xu2022document} and faithfulness \citep{durmus2020feqa}.  We
would also like to parametrize the oracle distribution and estimate
it from data, so as to derive even more accurate sentence labels.

\bibliography{custom}
\bibliographystyle{iclr2023_conference}

\newpage

\appendix
\section{Greedy Search Algorithm}
\label{appendix:greedy}
\begin{figure}[ht]
\centering
\begin{minipage}{0.7\textwidth}
\input{algo_greedy}
\end{minipage}
\end{figure}

\section{Label Statistics}
\label{appendx:beam}

\begin{tabular}{m{6cm}m{6cm}}
\begin{minipage}{\linewidth}
\pgfplotstableread{
Rank Ratio
1       90.37929228697539
2       5.588389316974639
3       1.7879853370240144
4       0.7181865788883071
5       0.3216877384603875
6       0.23191441609934915
7       0.19450886511558316
8       0.09725443255779158
9       0.059848881574025584
10      0.07481110196753198
}\loadedcnndm

\pgfplotstableread{
Rank Ratio
1       79.45570971184632
2       4.749199573105656
3       2.792600498043401
4       1.1739594450373532
5       1.0672358591248665
6       0.4091070793311989
7       0.5158306652436855
8       0.5691924581999288
9       0.4091070793311989
10      0.19565990750622556
}\loadedmultinews

\pgfplotstableread{
Rank Ratio
1       91.98473282442748
2       5.228120007101012
3       1.4290786437067282
4       0.5769572164033375
5       0.2751642109000533
6       0.21303035682584767
7       0.14202023788389845
8       0.062133854074205574
9       0.02662879460323096
10      0.017752529735487306
}\loadedxsum

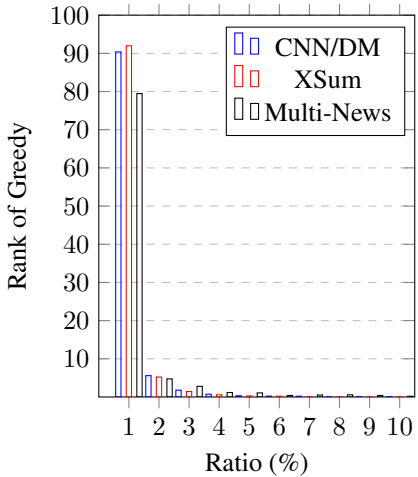
\begin{figure}[H]
\centering
\begin{tikzpicture}
\begin{axis}[
    ybar,
    bar width=2pt,
    width=0.7\columnwidth,
    height=2in,
    scale only axis,
    ylabel = Rank of Greedy,
    xlabel = Ratio (\%),
    xmin=0, xmax=10.5,
    ymin=0, ymax=100,
    xtick={1,2,3,4,5,6,7,8,9,10},
    ytick={10,20,30,40,50,60,70,80,90,100},
    legend pos=north east,
    ymajorgrids=true,
    grid style=dashed,
]

\addplot[
    draw=blue,
    ]
    table[x=Rank, y=Ratio] {\loadedcnndm};
    \addlegendentry{CNN/DM}

\addplot[
    draw=red,
    ]
    table[x=Rank, y=Ratio] {\loadedxsum};
    \addlegendentry{XSum}
    
\addplot[
    draw=black,
    ]
    table[x=Rank, y=Ratio]{\loadedmultinews};
    \addlegendentry{Multi-News}

\end{axis}
\pgfresetboundingbox
\path
  (current axis.south west) -- ++(-0.4in,-0.4in)
  rectangle (current axis.north east) -- ++(0.1in,0.1in);
\end{tikzpicture}
\caption{\label{fig:greedy_rank} 
Distribution of greedy oracles over top beams across three validation sets. 
} 
\end{figure}
\end{minipage}
& 
\begin{minipage}{\linewidth}
\pgfplotstableread{
x   y
1       99.70
2       99.90
4       99.97
8       99.98
16      99.99
32      100.00
64      100.00
128     100.00
256     100.00
}\loadmultinews

\pgfplotstableread{
x   y
1       99.7595
2       99.9779
4       99.9977
8       99.9998
16      99.9998
32      99.9998
64      99.9998
128     100.0000
256     100.0000

}\loadcnn

\pgfplotstableread{
x   y
1       99.6530
2       99.9423
4       99.9969
8       99.9999
16      100.0000
32      100.0000
64      100.0000
128     100.0000
256     100.0000

}\loadxsum

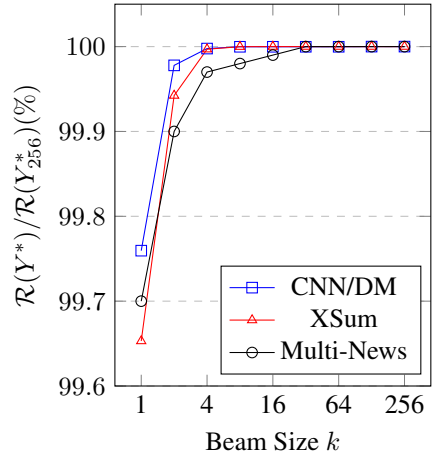
\begin{figure}[H]
\centering
\begin{tikzpicture}
\begin{axis}[
    width=0.7\columnwidth,
    height=2in,
    scale only axis,
    xlabel = Beam Size $k$,
    ylabel = $\mathcal{R}(Y^\ast)/\mathcal{R}(Y^\ast_{256}) (\%)$,
    ymin=99.6, ymax=100.05,
    symbolic x coords={1,2,4,8,16,32,64,128,256},
    ytick={99.5, 99.6, 99.7, 99.8, 99.9, 100.0},
    legend pos=south east,
    ymajorgrids=true,
    grid style=dashed,
]

\addplot[
    color=blue,
    draw=blue,
    mark=square,
    ]
    table[x=x, y=y] {\loadcnn};
    \addlegendentry{CNN/DM}

\addplot[
    color=red,
    draw=red,
    mark=triangle,
    ]
    table[x=x, y=y] {\loadxsum};
    \addlegendentry{XSum}
    
\addplot[
    color=black,
    draw=black,
    mark=o,
    ]
    table[x=x, y=y] {\loadmultinews};
    \addlegendentry{Multi-News}

\end{axis}

\end{tikzpicture}

\caption{\label{fig:best} 
Relative quality of best beams for beam labeling; results shown
different beam sizes across three validation sets.  }
\end{figure}
\end{minipage}
\end{tabular}

\textbf{Quality of Greedy Labels} Figure~\ref{fig:greedy_rank} shows
the distribution of greedy oracles over position in beam search
results, as ranked by $\mathcal{R}$ ($k=256$).
As we can see, around $8\%-20\%$ greedy labels are not top-ranked, and
can therefore be potentially improved with beam labels. However, as
shown in our experimental results, this improvement does not lead to a
summarization system that is consistently better across different
tasks.

\textbf{Effects of Beam Size} As beam search does not guarantee a global
optimum either, we further calculate
$\mathcal{R}(Y^\ast)/\mathcal{R}(Y^\ast_{256})$ to evaluate the
\textit{relative} quality of the top beam $Y^\ast$ (the top beam found
by varied beam sizes), compared against $Y^\ast_{256}$ (the top beam
found by beam size 256).  Figure \ref{fig:best} shows that the quality
of the top beam converges when beam size increases to $64$.  However,
as oracles are not necessarily indicative of actual model performance
(see Section \ref{sec:further} for details), we view beam size as a
hyperparameter for optimization.

\section{Equivalence Proof}
\label{appendix:proof}
Given  input document~$D$, sentence-level inference of a
non-autoregressive summarization model~$\theta$ is conditionally
independent, and the likelihood of an oracle summary
$Y^\ast=\{y^\ast_i\}_{i=1}^m$ (in its multi-hot representation over
$m$~document sentences) is calculated as:
\begin{equation}
    \label{eq:cond_ind}
    p_\theta({Y}^\ast | D) = \prod_{i=1}^m p_\theta(x_i=y^\ast_i | D).
\end{equation}

In this case, we show that maximizing the oracle expectation for all
sentences is equivalent to the objective in Equation~(\ref{eq:obj}):
\begin{align}
\max
    \prod_{i=1}^m p_\theta(x_i={\ell}'_i|D, S) 
    &= \prod_{i=1}^m
    \EXP_{Y^\ast \sim p(Y^\ast|D,S)}
    \left[ 
    \mathcal{R}(Y^\ast, S)
    p_\theta(x_i | Y^\ast, D)
    \right]
    &&\textcolor{gred}{\rhd \text{ Plug in \textsc{Oreo}}} \nonumber
    \\
    &= \EXP_{Y^\ast \sim p(Y^\ast|D, S)}
    \left[ 
    \mathcal{R}(Y^\ast, S)
    \prod_{i=1}^m p_\theta(x_i = y^\ast_i | D)
    \right]
    &&\textcolor{gred}{\rhd \text{ Take out $\EXP$}} \nonumber
    \\
    &= \EXP_{Y^\ast \sim p(Y^\ast|D, S)}
    \left[ 
    \mathcal{R}(Y^\ast, S)
    p_\theta (Y^\ast | D) 
    \right].
    &&\textcolor{gred}{\rhd \text{ Apply Eq. (\ref{eq:cond_ind})}}
\end{align}

We note that our objective in Equation (\ref{eq:obj}) serves as a
lower bound of the classic extractive summarization objective $\max
p_\theta (Y^\ast|D)$ \textit{weighted} by oracle evaluation:
\begin{align}
&\max \EXP_{Y^\ast \sim p(Y^\ast|D, S)}
    \left[ 
    \mathcal{R}(Y^\ast, S)
    p_\theta (Y^\ast | D) 
    \right]\\
&=\sum^{\Y}_{Y^\ast}
p(Y^\ast | D, S)
\mathcal{R}(Y^\ast, S)
p_\theta (Y^\ast | D)\\
&\le \mathcal{R}(Y^\ast _{\text{best}}, S)
p_\theta (Y^\ast_{\text{best}} | D)
\text{ where }
{Y}^\ast_{\text{best}} = \argmax_{Y^\ast\in \Y} \mathcal{R}(Y^\ast,S)\\
&\propto p_\theta (Y^\ast_{\text{best}} | D)
\end{align}
The equality holds only if the oracle distribution $p(Y^\ast|D, S)$ is a Dirac delta distribution  
$\delta(Y^\ast-Y^\ast_{\text{best}})$.

\section{Effects of Oracle  Distribution}
\label{appendix:oracle_dist}

\begin{table}[H]
\centering
\def\arraystretch{1.2}

\caption{\label{tab:oracle_dist}
 \textsc{Oreo} results with different oracle distributions on {CNN/DM} validation set.}

\begin{tabular}{lccc}
\thickhline
\textbf{Systems}
& \textsc{Rouge-1} & \textsc{Rouge-2} & \textsc{Rouge-L}\\
\hline
\multicolumn{4}{l}{\textsc{BertSum}}
\\
~~~~Greedy
&44.00 &20.73 &40.45
\\
~~~~\textsc{Oreo}, $U(1,t)$
&\textbf{44.26} &\textbf{20.92} 
&\textbf{40.69}\\
\hline
~~~~\textsc{Oreo}, $A_r(1,16)$
&44.06 &20.76 &40.50
\\
~~~~\textsc{Oreo}, $A_q(1,16)$
&44.17 &20.83 &40.60
\\
~~~~\textsc{Oreo}, $A_{\ell}(1,16)$
&44.03 &20.73 &40.42
\\
~~~~\textsc{Oreo}, $A_{p}(1,16)$
&43.91 &20.45 &40.26
\\
\thickhline
\end{tabular}

\end{table}

We devise and experiment with several oracle distributions that assign non-uniform probability to top~$t$ beams:

\begin{enumerate}
    \item \textbf{Annealing over Rank $A_r$} decreases the
      unnormalized weight from 1 to 0 over top beams, assuming that
      the oracle distribution positively correlates with
      \textit{hypothesis rank}.
    \item \textbf{Annealing over Quality $A_q$} sets the unnormalized
      weight for a top beam $Y^\ast$ as $\mathcal R(Y^\ast)$, assuming
      that the oracle distribution positively correlates with
      \textit{hypothesis evaluation score}.
    \item \textbf{Annealing over Locality $A_{\ell}$}
    defines the locality of a hypothesis $Y$ to be proportional to mean of  sentence-level scores $\overline{\mathcal{R}(y^\ast_j, S)}$, $y^\ast_j \in Y^\ast$. 
    This is based on the assumption that a hypothesis is more
    \textit{local} if its sentences are, by themselves, high-scoring.
    Hypothetically, these sentences stand a higher chance to be
    selected by a non-autoregressive sequence labeling model which
    presumably focuses more on their individual features rather than
    collective information \citep{matchsum}.
    \item \textbf{Annealing over Position Rank $A_p$} decreases the
      unnormalized weight from 1 to 0 over top beams which are
      reversely ranked by their position in the original document,
      assuming that oracle distribution positively correlates with
      \textit{document position}.
\end{enumerate}


 Table \ref{tab:oracle_dist} presents our results on CNN/DM validation set.  As we
 can see, the above-mentioned distributions do not yield better
 results than simply adopting a uniform distribution (third row). We
 believe this is because hand-crafted distributions are all associated
 with strong assumptions, which may not be valid for real-world
 summarization data.  However, we note that most of these
 distributions still manage to outperform greedy labels, showing
 consistent gains when considering information from multiple oracles.
 
 Apart from heuristic oracle distributions, we could also learn a
 parametrized distribution from data. For instance, a model with a
 uniform oracle distribution could be trained to derive a potentially
 more accurate estimation from its predictions.  A new set of sentence
 labels would be then calculated with Equation~(\ref{eq:final_obj}),
and used to improve the optimization of a new model.


\section{Details of Experimental Settings}
\label{appendx:setting}

\begin{table}[H]
\centering
\def\arraystretch{1.2}

\caption{Hyperparameters for supervised training of \textsc{BertSum}
  on five summarization datasets.}

\begin{tabular}{lrrrrr}
\thickhline
Monolingual & CNN/DM & XSum & Multi-News  & Reddit & WikiHow\\  \thickhline
Beam size $k$ & 256 & 16 & 16 & 256 & 16\\
Oracle distribution $t$
&16
&16
&16 
&32
&16\\
\thickhline
\end{tabular}
\label{tab:hyperparams}
\end{table}

\paragraph{Monolingual Extractive Summarization}
We used three GeForce RTX 2080 GPUs for model training and
\texttt{bert.base} in our experiments.  We refer interested readers to
\citet{bertsum} for detailed training configurations which are
identical to ours. Following \citet{bertsum}, we used the Python
package \texttt{pyrouge} for calculating \textsc{Rouge}.  For our the
proposed labeling methods, we searched over the following $(k, t)$
pairs: $(256, 32), (256, 16), (256, 8), (32, 32), (16, 16), (8, 8)$.
We show the best-performing hyperparameter combinations for each
dataset in Table~\ref{tab:hyperparams}.  We used standard
parameter settings for all experiments: {ROUGE-1.5.5.pl -c 95 -m -r
  1000 -n 2 -a}.  We used the datasets as preprocessed by
\citet{matchsum} which can be accessed at:
\url{https://github.com/maszhongming/matchsum}.

\paragraph{Cross-Lingual Extractive Summarization}
In our cross-lingual experiments we used four GeForce RTX 2080 GPUs
for model training with \texttt{xlmr.base} and \texttt{mbert.base}.
Particularly, interval embeddings \citep{bertsum} were used in
\textsc{MBertSum} but not in \textsc{XlS} since XLM-R removes
segment embeddings from the input following RoBERTA \citep{roberta}.
We refer readers to \citet{jia2022neural} for details on training
configuration; we made minimal adjustments to adapt to our training
environment, i.e., no training hyperparameters were specifically
optimized for our method.  We set the batch size to 4, and accumulated
gradients every 32~steps.  Following \citet{jia2022neural}, we used
word replacement rate of~0.5 to learn cross-lingual representation
alignment.  We fine-tuned models on the English data with a learning
rate of $2 \times 10^{-3}$ for 50,000 optimization steps, and a
warm-step of~10,000.  Following \citet{jia2022neural}, we used the
Python package \texttt{spacy} for non-English hypothesis/reference
tokenization, and \texttt{pyrouge} for ROUGE calculation.

\paragraph{Abstractive Summarization}
In our abstractive summarization experiments we used four GeForce RTX
2080 GPUs for model training with \texttt{bart.large}; the latter was
also used in our baseline \textsc{Bart} system and to initialize
\textsc{GSum}.  Due to GPU memory limitations, we set the maximum
length of an input document to 640 tokens (with the excess clipped)
and used half float precision for efficient training.  We used one
sample for each GPU, and accumulated gradients every 32 steps.  We
fine-tuned all models on CNN/Daily Mail with a learning rate of $3
\times 10^{-5}$ for 20,000 optimization steps, and a warm-step of 500.
Following the evaluation steps in BART \citep{bart} and \textsc{GSum}
\citep{dou2020gsum}, we used
\texttt{file2rouge}\footnote{\url{https://github.com/pltrdy/files2rouge}}
to evaluate abstractive summaries.


\section{Extended  Results}
\label{appendix:more_results}
\begin{table}[H]
\def\arraystretch{1.2}
\bgroup

\caption{\label{tab:result_xl_mbert} 
Cross-lingual zero-shot performance on test sets of \textbf{MLSum} in \textsc{Rouge-L}.}

\centerline{
\begin{tabular}{lcccccc}
\thickhline
{\textbf{Systems}} 
& \textbf{De} & \textbf{Es} & \textbf{Fr} & \textbf{Ru} & \textbf{Tr} & \textbf{AVG}
\\
\hline
{\sc MBertSum}
\\
~~~~Greedy
&\textit{22.68}	&\textit{20.44}	&\textit{22.70}	&\textit{8.71}	&\textit{27.89}	&\textit{20.48}
\\
~~~~{Beam}
&28.36	&{20.55}	&{22.74}	&\textbf{9.30}	&29.38 & 22.07
\\
~~~~\textsc{Oreo}
&\textbf{29.13}	&\textbf{20.62}	&\textbf{22.82}	&9.13	&\textbf{30.78}	&\textbf{22.50}
\\
\thickhline
\end{tabular}
}
\egroup
\end{table}

 \paragraph{Cross-Lingual Results}
 We further initialize \textsc{BertSum} with mBERT \citep{devlin2019bert}.
 As we can in Table~\ref{tab:result_xl_mbert}, \textsc{mBertSum} 
 finetuned with greedy labels shows inferior performance across
 languages. Nevertheless, \textsc{Oreo} leads to substantial
 performance gains on both German and Turkish (we observe a similar
 trend when \textsc{BertSum} initialized with XLM-R).

\begin{table}[H]
\centering
\def\arraystretch{1.2}

\caption{\label{tab:results_abs_more}
Results for abstractive summarization on \textbf{CNN/DM} test set. 
}

\begin{tabular}{lccc}
\thickhline
\textbf{Systems}
& \textsc{Rouge-1} & \textsc{Rouge-2} & \textsc{Rouge-L}\\
\hline
\multicolumn{4}{l}{\textsc{GSum} trained with greedy \textsc{Oracle} guidance} 
\\
~~~~Greedy
& \textit{44.40} & \textit{21.52} & \textit{41.23} \\
~~~~Beam
& 44.41 & 21.55 & 41.26 \\
~~~~\textsc{Oreo}
& \textbf{44.48} & \textbf{21.60} & \textbf{41.33}\\
\hdashline
\multicolumn{4}{l}{\textsc{GSum}
trained with \textsc{Oreo Oracle} guidance
}\\
~~~~Greedy
&44.66
&21.72
&41.43
\\
~~~~Beam
&44.68
&21.74
&41.47
\\
~~~~\textsc{Oreo}
&\textbf{44.81}
&\textbf{21.83}
&\textbf{41.60}
\\
\thickhline
\end{tabular}

\end{table}

\paragraph{Abstractive Results}
We show extended abstractive results in Table
\ref{tab:results_abs_more}.  The first block shows performance of
vanilla \textsc{GSum} \citep{dou2020gsum} which uses greedy extractive
oracles as guidance during training.  During inference, we compare to
three types of extractive guidance produced by \textsc{BertSum}
trained with greedy, beam and \textsc{Oreo} labels.  Despite the
training-testing discrepancy in guidance, extractive guidance with
\textsc{Oreo} labels helps generate better downstream abstractive
summaries.

To further validate the effectiveness of \textsc{Oreo} on the
optimization of \textsc{GSum}, we trained \textsc{GSum} with
\textsc{Oreo Oracle}, and the results are shown in the second block.
As we can see, adopting \textsc{Oreo} guidance during training further
boosts system performance, while the system using \textsc{Oreo} for
both training and testing achieves the best results, i.e., 0.37
\textsc{Rouge-L} improvement over \textsc{GSum}.



\section{System Output}
\label{appendix:sys_out}
\begin{table}[H]
\centering
\begin{tabular}{p{13.6cm}}\\
\thickhline \textbf{Document}: The largest single high-definition map
of mysterious dark matter has been produced.  \textcolor{gblue}{It is
  the first in a series of maps of the cosmos that will eventually
  allow a 3D view of dark matter across one eighth of the night sky.}
And the map should allow astronomers to study how galaxies formed in
the universe.  University of Manchester researchers have revealed an
HD dark matter map (shown). It shows clumps of mystery particles
across 0.4 per cent of the sky.  \textcolor{gred}{The goal is to
  eventually map 12.5 per cent over five years.}  Red here shows more
dark matter, and blue shows less. The moon is shown top left for
scale. A team from the University of Manchester, led by Dr Sarah
Bridle, has spent the past two years measuring the shapes of galaxies
used to construct the map. And the map was released today at the April
meeting of the American Physical Society in Baltimore, Maryland.  It
was created using data captured by the Dark Energy Camera, a
570-megapixel imaging device that is the primary instrument for the
Dark Energy Survey (DES) led by Fermilab in the US...
\\
\\
\textbf{Reference Summary}: 
University of Manchester researchers reveal HD dark matter map. 
It shows clumps of mystery particles across 0.4 per cent of the sky. 
\textcolor{gred}{The goal is to eventually map 12.5 per cent over five years.}
Results could help reveal how galaxies form in the universe.
\\
\hline

\multicolumn{1}{c}{} \\ 
\multicolumn{1}{c}{Extractive Summarization with \textsc{BertSum}}\\   
\hline

\textbf{Greedy}: 
University of Manchester researchers have revealed an HD dark matter map (shown).
It shows clumps of mystery particles across 0.4 per cent of the sky.
\textcolor{gblue}{It is the first in a series of maps of the cosmos that will eventually allow a 3d view of dark matter across one eighth of the night sky.}
\\\\
\textbf{Beam}: 
University of Manchester researchers have revealed an HD dark matter map (shown).
\textcolor{gred}{The goal is to eventually map 12.5 per cent over five years.}
\textcolor{gblue}{It is the first in a series of maps of the cosmos that will eventually allow a 3d view of dark matter across one eighth of the night sky.}
\\
\\
\textbf{\textsc{Oreo}}: 
University of Manchester researchers have revealed an HD dark matter map (shown).
It shows clumps of mystery particles across 0.4 per cent of the sky.
\textcolor{gred}{The goal is to eventually map 12.5 per cent over five years.}
\\

\hline                                 
\multicolumn{1}{c}{} \\ 
\multicolumn{1}{c}{Abstractive Summarization with \textsc{GSum}}\\
\hline

\textbf{Greedy}:
University of Manchester researchers have revealed an HD dark matter map. It shows clumps of mystery particles across 0.4 per cent of the sky. 
\textcolor{gred}{The goal is to eventually map 12.5 per cent over five years.} 
\textcolor{gblue}{It is the first in a series of maps of the cosmos that will eventually allow a 3D view of dark matter across one eighth of the night sky.} 
And the map should allow astronomers to study how galaxies formed in the universe.\\\\
\textbf{Beam}:
University of Manchester researchers have revealed an HD dark matter map. It shows clumps of mystery particles across 0.4 per cent of the sky. 
\textcolor{gred}{The goal is to eventually map 12.5 per cent over five years.}
\textcolor{gblue}{It is the first in a series of maps of the cosmos that will eventually allow a 3D view of dark matter across one eighth of the night sky.} 
And the map should allow astronomers to study how galaxies formed in the universe.\\\\
\textbf{\textsc{Oreo}}:
University of Manchester researchers have revealed an HD dark matter map of the cosmos. 
It shows clumps of mystery particles across 0.4 per cent of the sky. 
Red here shows more dark matter, and blue shows less. \textcolor{gred}{The goal is to eventually map 12.5 per cent over five years.}
And the map should allow astronomers to study how galaxies formed in the universe.\\
\thickhline
\end{tabular}

\caption{\label{tab:summ_types} 
Examples of system output on the CNN/DM test set. 
We illustrate differences among labeling algorithms with a sentence
from the reference summary labeled in \textcolor{gred}{red}.
\textsc{BertSum} trained with {\textsc{Oreo}} labels includes
the sentence in its extract.  In contrast, {Greedy} selects a
suboptimal, verbose sentence highlighted in \textcolor{gblue}{blue},
potentially due to its position in the beginning of the original
document (lead bias).  The {Beam} extract includes both sentences and
is therefore most redundant.  Using these extracts as inference
guidance, \textsc{GSum} creates abstractive summaries which for
{Greedy} and {Beam} are identically verbose, while
{\textsc{Oreo}} summary is more concise. }

\end{table}

\section{Bound Preservation}
\label{appendix:bound_preservation}
\begin{figure}[H]
    \centering
  \includegraphics[width=\textwidth]{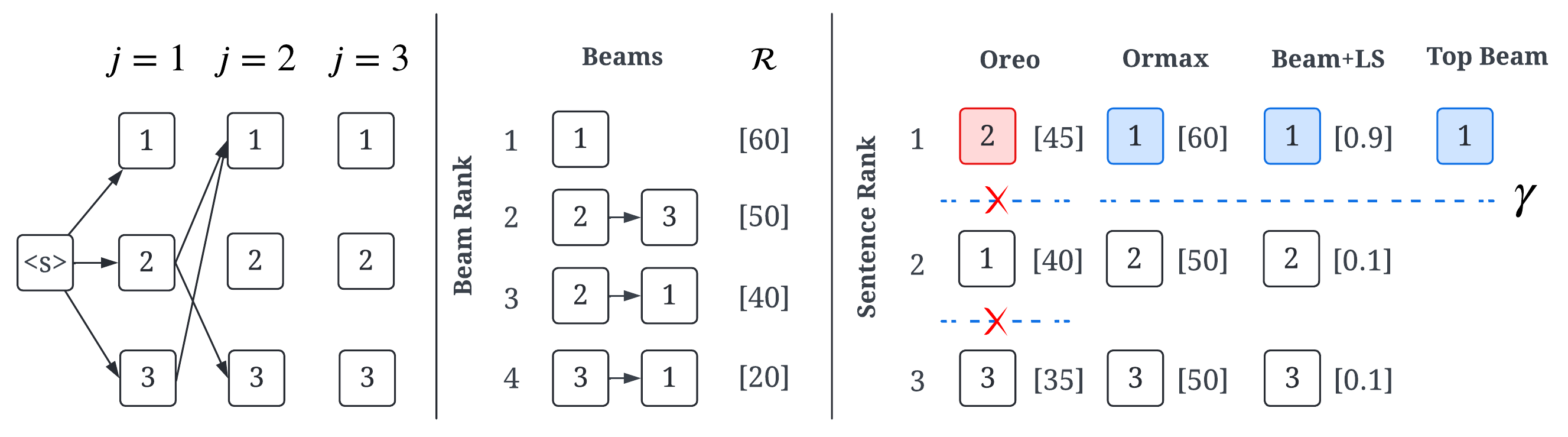}
  \caption{\label{fig:eps} Illustration of beam search (left; early
    stopped at $j=2$), ranked beams (middle), and ranked sentences
    (right). We show sentences (in squares) and their scores (in
    brackets) under different labeling algorithms.  For \textsc{Oreo},
    there does not exist a~$\gamma$ that halves the ranked list in a
    way that the top half is identical to the sentences selected by
    the top beam.  }
\end{figure}

\end{document}